\documentclass{article}

\usepackage{microtype}
\usepackage{graphicx}
\usepackage{subfigure}
\usepackage{booktabs} 

\usepackage{hyperref}
\usepackage{enumitem,kantlipsum}
\usepackage{listings}
\usepackage{tcolorbox}
\usepackage{xcolor}

\usepackage[accepted]{icml2024}

\usepackage{amsmath}
\usepackage{amssymb}
\usepackage{mathtools}
\usepackage{amsthm}
\usepackage{multirow}

\usepackage[capitalize,noabbrev]{cleveref}

\theoremstyle{plain}

\theoremstyle{definition}

\theoremstyle{remark}

\usepackage{amssymb}
\usepackage{pifont}
\usepackage[textsize=tiny]{todonotes}

\definecolor{codegreen}{rgb}{0,0.6,0}
\definecolor{codegray}{rgb}{0.5,0.5,0.5}
\definecolor{codepurple}{rgb}{0.58,0,0.82}
\definecolor{backcolour}{rgb}{0.95,0.95,0.92}

\lstdefinestyle{mystyle}{
    backgroundcolor=\color{backcolour},   
    commentstyle=\color{codegreen},
    keywordstyle=\color{magenta},
    numberstyle=\tiny\color{codegray},
    stringstyle=\color{codepurple},
    basicstyle=\ttfamily\footnotesize,
    breakatwhitespace=false,         
    breaklines=true,                 
    captionpos=b,                    
    keepspaces=true,                 
    numbers=left,                    
    numbersep=5pt,                  
    showspaces=false,                
    showstringspaces=false,
    showtabs=false,                  
    tabsize=2
}

\icmltitlerunning{DFA-RAG: Conversational Semantic Router for Large Language Model with Definite Finite Automaton}

\begin{document}

\twocolumn[
\icmltitle{DFA-RAG: Conversational Semantic Router for Large Language Model \\ with Definite Finite Automaton}



\icmlsetsymbol{equal}{*}

\begin{icmlauthorlist}
\icmlauthor{Yiyou Sun}{wisc,nec}
\icmlauthor{Junjie Hu}{wisc}
\icmlauthor{Wei Cheng}{nec}
\icmlauthor{Haifeng Chen}{nec}
\end{icmlauthorlist}

\icmlaffiliation{wisc}{Department of Computer Science, University of Wisconsin}
\icmlaffiliation{nec}{NEC Laboratories America, inc., Princeton, USA}

\icmlcorrespondingauthor{Yiyou Sun}{sunyiyou@nec-labs.com}
\icmlcorrespondingauthor{Haifeng Chen}{haifeng@nec-labs.com}

\icmlkeywords{Machine Learning, ICML}

\vskip 0.3in
]



\printAffiliationsAndNotice{\icmlEqualContribution} 

\begin{abstract}
This paper introduces the retrieval-augmented large language model with Definite Finite Automaton (DFA-RAG), a novel framework designed to enhance the capabilities of conversational agents using large language models (LLMs). Traditional LLMs face challenges in generating regulated and compliant responses in special scenarios with predetermined response guidelines, like emotional support and customer service. Our framework addresses these challenges by embedding a Definite Finite Automaton (DFA), learned from training dialogues, within the LLM. This structured approach acts as a semantic router which enables the LLM to adhere to a deterministic response pathway. 
The routing is achieved by the retrieval-augmentation generation (RAG) strategy, which carefully selects dialogue examples aligned with the current conversational context. 
The advantages of DFA-RAG include an interpretable structure through human-readable DFA, context-aware retrieval for responses in conversations, and plug-and-play compatibility with existing LLMs. 
Extensive benchmarks validate DFA-RAG's effectiveness, indicating its potential as a valuable contribution to the conversational agent.

\end{abstract}

\section{Introduction}
\label{sec:intro}

Recent advancements in machine learning, particularly in large language models (LLMs), have provided more possibilities in various fields. Their applications range from document completion to chatbots~\cite{ouyang2022training,openai2023gpt4}. These conversational agents with LLMs are remarkably adaptable and have proven effective in diverse settings including customer service, marketing, education, and healthcare~\cite{wang2023chatcad}.
Compared to traditional, rule-based dialogue systems ~\cite{abdul2015survey,hussain2019survey,adamopoulou2020chatbots}, LLMs offer greater efficiency, scalability, and dynamism~\cite{medeiros2019towards}. However, their application in practical scenarios often necessitates adherence to specific workflows or policies. For instance, an Emotional Support Chatbot ~\cite{medeiros2019towards} must tailor its responses to various stress conditions, while customer service bots typically follow predetermined response guidelines.

\begin{figure*}[htb]
    \centering
    \includegraphics[width=\linewidth]{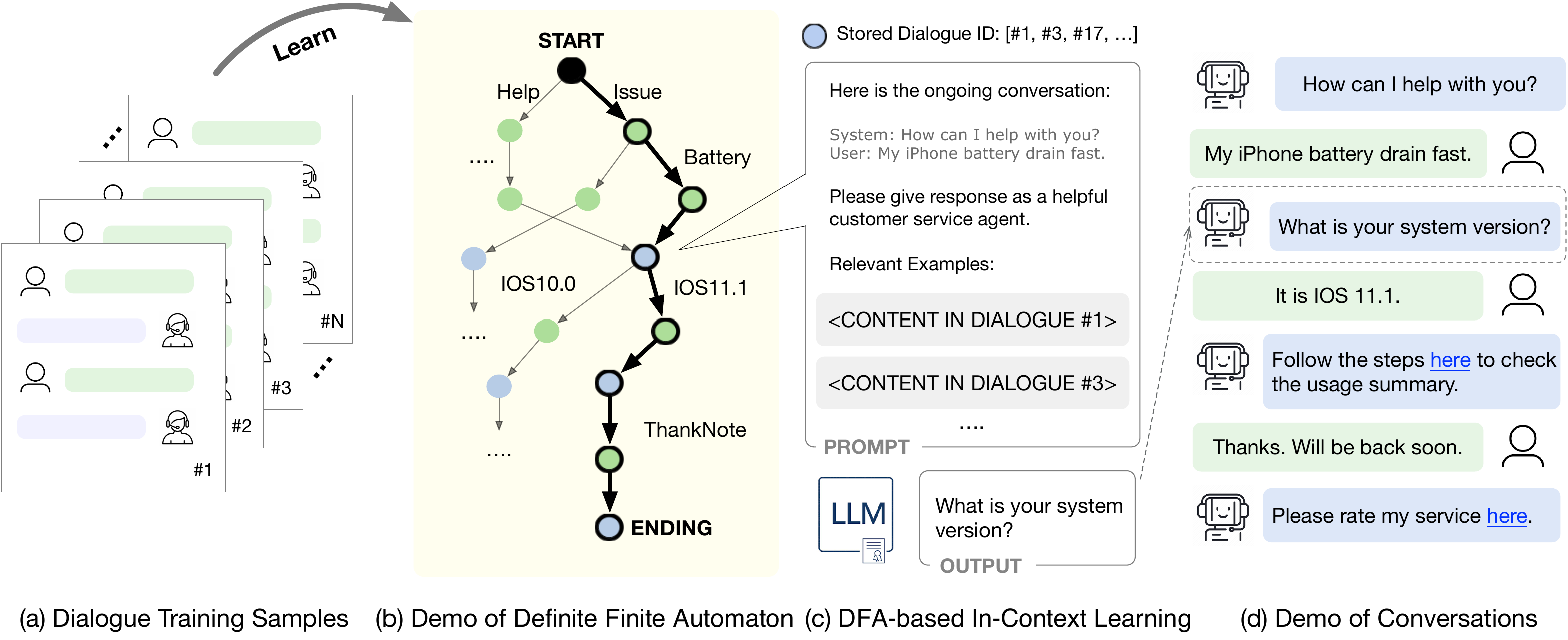}
    \vspace{-0.4cm}
    \caption{Illustration of the DFA-RAG Framework. (a) shows the training set with dialogues. (b) demonstrates the Definite Finite Automaton (DFA) which represents the workflow learned from the dialogues. Blue and green dots represent the states of the user and system respectively. The states are transited by keywords in conversations.   (c) outlines the DFA-based In-Context Learning process, where the LLM is guided by the DFA to provide contextually relevant responses. (d) showcases sample conversations between a user and the LLM.}
    \label{fig:teaser}
\end{figure*}

In these scenarios, popular LLMs like GPT-3.5 or GPT-4 may generate inappropriate or misleading content~\cite{wang2023decodingtrust} without specialized optimization. Fine-tuning these models, while effective, is not always viable due to open-source requirements, the need for intricate design~\cite{ding2023parameter}, and substantial data to mitigate overfitting risks~\cite{OverfittingRisksLLMsNCCGroup}. An alternative method, Retrieval-Augmented Generation (RAG)~\cite{lewis2020rag}, which references an additional knowledge base in generating a response, has been explored to address these issues. Since the generation quality of RAG is known to be sensitive to the sample selection~\cite{gao2023retrieval}, it is therefore critical to design an effective selection strategy that ensures the retrieval of the most relevant and contextually appropriate samples. This poses unique challenges to apply RAG in conversational scenarios, as one must identify partial conversational flows within historical training dialogues that closely match the current conversation context.

Targeting these challenges, we formally introduce the new framework, DFA-based retrieval-augmented generation (dubbed \textbf{DFA-RAG}). This framework assumes that a specific workflow is embedded within the training data dialogues, which we model using a Definite Finite Automaton (as illustrated in Figure~\ref{fig:teaser}). Acting as a Semantic Router~\cite{horsey2024semantic} like a decision-making layer, DFA-RAG routes conversations through a predefined trajectory. This ensures that the LLM adheres to the workflow encoded in the DFA. Specifically, each conversational utterance corresponds to a particular DFA state, where each state encapsulates responses from similar historical contexts. By utilizing these historical examples as retrieved samples, DFA-RAG effectively guides the LLM to follow a deterministic DFA-based response pathway. The DFA-RAG framework offers several compelling advantages:
\begin{enumerate}[leftmargin=*]
    \item \textbf{Interpretable Structure}: Unlike RAG (Retrieval-Augmented Generation)~\cite{lewis2020rag} which fuses information on uninterpretable embedding spaces, our approach ensures responses are generated by human-readable structures (DFA), enhancing the trustworthiness of LLM-based chatbots.
   \item \textbf{Context-Aware Retrieval}: Unlike most retrieval-based strategies that treat dialogues as undivided wholes, our approach divides each dialogue into manageable segments. This granular approach allows for more precise and contextually relevant information retrieval, enhancing the chatbot's ability to respond accurately and appropriately to various conversational nuances.
   
   \item \textbf{Plug-and-Play Compatibility}: The DFA, derived from training data, is designed for easy integration with any pre-trained LLM. This flexibility allows for widespread application across different LLM platforms, making DFA-RAG a versatile tool in conversational AI.

   \item \textbf{Strong Empirical Performance}: In domain-specific conversations, DFA-RAG has shown superior performance, as demonstrated by extensive evaluation benchmarks. This empirical evidence underscores the effectiveness of the proposed framework in delivering high-quality, domain-relevant responses.

\end{enumerate}


\section{Problem Setup}
\label{sec:prelim}


In our setting, we assume access to a sample set of dialogues in the application domain of interest, such as customer service or emotional support. This dataset may be pre-collected through interactions between a human agent and a customer or patient. We now formally describe the data setup.

\textbf{Data Setup.} We consider the training set with $N$ dialogues $\mathcal{D}_{\text{tr}} = \{x_i\}^{N}_{i=1}$. Each dialogue is a sequence of utterances, denoted as $x_i = (z_1, z_2, \ldots, z_{l_i})$, where $l_i$ represents the length of the $i$-th dialogue. We assume the utterances alternate between the agent and the user. 

\textbf{Goal.} At the inference stage, the LLM-based agent generates the next utterance based on the context of an incomplete dialogue $\bar{x} = (z_1, z_2, \ldots, z_j)$:
    $$\hat{z}_{j+1} \sim \mathbb{P}(\hat{z} | \bar{x}, \mathcal{D}_{\text{tr}}). $$
The objective is for the output $\hat{z}_{j+1}$ to closely match the human agent's response $z_{j+1}$.

\section{Methodology}
\label{sec:method}



Human agents often provide responses that involve navigating a pre-determined landscape of options, heuristics, and potential outcomes. This process can be conceptualized as exploring a special branch of a network of interconnected paths and junctions. 
In contrast, current LLMs like GPT-4 lack the intrinsic ability to mimic this human-like pipeline-guided behavior in response generation.
This limitation becomes particularly pronounced in scenarios where adherence to specific workflows is paramount, such as in customer service or healthcare settings. In these cases, traditional LLMs might generate responses that are misaligned~\cite{wang2023decodingtrust} with the desired conversational trajectory or decision pathway.

To address this gap, our methodology with DFA-RAG embeds a structured pipeline within the operational paradigm of the LLM. By integrating a DFA, we propose to direct the LLM's response generation process, aligning it more closely with the decision paths of typical human agents. In Section~\ref{sec:model_dfa}, we introduce how we can model the conversations as a DFA. 
Later in Section~\ref{sec:learn_dfa}, we delve into details of constructing the DFA from training dialogue datasets. 
Subsequently, in Section~\ref{sec:dfa_guide}, we illustrate the mechanism by which the learned DFA guides the LLM. 

\subsection{Modeling Conversations with DFA}
\label{sec:model_dfa}

\subsubsection{Preliminaries of DFA}

Automata has a long history of study in theoretical computer science, linguistics, and other related fields~\cite{minsky1956some, kleene1956automata}. A deterministic finite automaton (DFA) can be specified as a tuple \((Q, \Sigma, \delta, q_0, F)\), where:
\begin{itemize} 
    \setlength\itemsep{-0.5em}
    \item \(Q\) is a finite set of states
    \item \(\Sigma\) is a finite input alphabet
    \item \(\delta: Q \times \Sigma \rightarrow Q\) is the transition function
    \item \(q_0 \in Q\) is the start state
    \item \(F \subseteq Q\) is the set of accept states
\end{itemize}
The DFA processes a string of symbols from \(\Sigma\) and changes its state according to the transition function \(\delta\). The string is accepted by the DFA if the automaton is in one of the accept states in \(F\) after processing all symbols. 

%

\begin{figure}[h]
    \centering
    \includegraphics[width=0.85\linewidth]{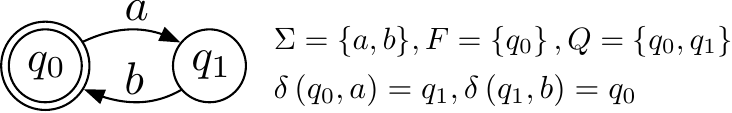}
    \vspace{-0.2cm}
    \caption{A demo of DFA recognizing string ``\textit{$(ab)^*$}".}
    \label{fig:dfa-demo}
\end{figure}

\subsubsection{Conversation as Tag-Sequence}
DFA are widely recognized for their capability to define alphabet sequences with a specific syntax. A classic example as shown in Figure~\ref{fig:dfa-demo} is a DFA with a two-state structure that can define strings like ``\textit{ababab}" and ``\textit{abab}". However, when it comes to modeling conversations, the challenge arises due to the complexity and variety of conversational elements. Unlike simple syntactic strings, the essence of a conversation lies predominantly at the semantic level. For instance, phrases such as ``\textit{My battery drains out fast}" and ``\textit{How come my phone can be only used for 1 hour?}" convey the same underlying issue in a customer service context, yet they do not share a single word in common. This disparity underscores the limitations of traditional DFA models in capturing the nuances of conversational semantics.

To address this, we assume each utterance in a conversation can be encapsulated by a set of ``keywords" or ``tags". (We use the term ``\texttt{tags}" in this paper). For example, the utterance ``How come my phone can be only used for 1 hour?" could be succinctly represented by the tag set $\{``\#\texttt{issues}", ``\#\texttt{battery}"\}$ in the customer service context. This method allows for the abstraction of utterances into a more manageable form, suitable for DFA modeling.

\subsubsection{Conversation Sets as DFA}

Building upon the idea of representing individual conversations as sequences of tags, we extend this concept to model entire sets of conversations using Deterministic Finite Automata (DFAs). Recall our earlier definition of a DFA as a tuple  \((Q, \Sigma, \delta, q_0, F)\). In this context, we adapt these components to fit our conversation modeling framework:
\begin{itemize}[leftmargin=*]
    \setlength\itemsep{-0.2em}
    \item \textbf{States }($Q$): Each state in the DFA represents a particular stage or context within a conversation. For instance, a state could represent the initiation of a conversation, a query about a specific issue, a response, or the conclusion of the interaction.
    \item \textbf{Alphabet} ($\Sigma$): The alphabet in this model comprises the set of all possible tags that we identified as representative of different utterances. These tags form the basic building blocks of our conversation sequences.
    \item \textbf{Transition Function} ($\delta$): The transition function maps a state and a tag to a subsequent state. It encapsulates the flow of conversation, determining how an utterance (through its tags) leads from one conversational context to another.
    \item \textbf{Start State} ($q_0$): The start state represents the beginning of a conversation. It could be a generic greeting or an initial query, setting the stage for the interaction.
    \item \textbf{Accept States} ($F$): These are the states indicating the completion of a conversation. An accept state could be reached after successfully addressing a query, reaching a satisfactory conclusion, or when the conversation naturally comes to an end.
\end{itemize}

\textbf{Remark on the dialogue tracking function $\mathcal{I}$.} In our DFA framework, a unique functionality is incorporated where each state maintains a record of the indices of dialogues that have traversed through it. This is accomplished by tracing the trajectory of tag sequences as the conversation unfolds. Formally, this tracking is represented by the function $\mathcal{I}(q)$, which maps a state $q$ to a set of dialogue IDs in 
$\{1, ..., N\}$. 
Crucially, it enables us to trace back and identify the most relevant dialogue samples that can be effectively utilized in in-context learning.

\begin{figure*}[htb]
    \centering
    \includegraphics[width=\linewidth]{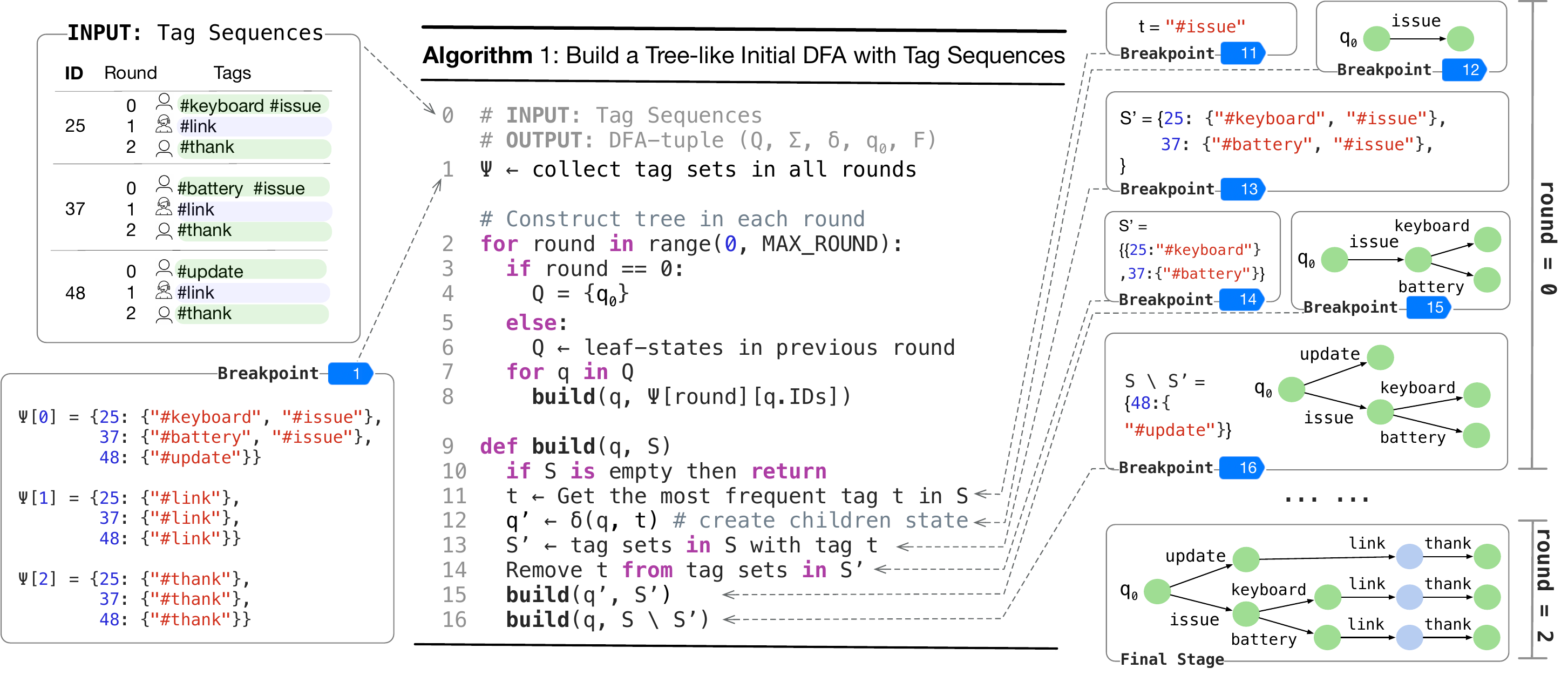}
    \vspace{-0.2cm}
    \caption{Illustration of the algorithm for building a tag tree, expressed in Python-style pseudocode. \textbf{(1)} Positioned \textbf{centrally} is the core algorithm, flanked on either side by demonstrative examples that ``print out" the algorithm's internal variables at various ``breakpoints".\textbf{ (2)} On the \textbf{left}, we begin by displaying the input data used in the demonstration. Following this, the first breakpoint, corresponding to \textit{Line 1} of the algorithm, showcases the variables storing the ID-ed tag sets across different conversational rounds. \textbf{(3) }
Shifting to the \textbf{right} side, we illustrate the evolution of the variables and the tree structures, starting from \textit{round 0} with the initial state $q_0$
. The breakpoints at \textit{Lines 15 $\&$ 16} forego stepping into the function, instead presenting the completed tree structures directly for clarity. To simplify the illustration, we omit the iterative process for \textit{rounds 1 $\&$ 2}, jumping straight to the final results.  }
    \label{fig:tree-alg}
\end{figure*}

\subsection{Learn DFA from Conversations}
\label{sec:learn_dfa}
\begin{figure}[htb]
    \centering
    \includegraphics[width=\linewidth]{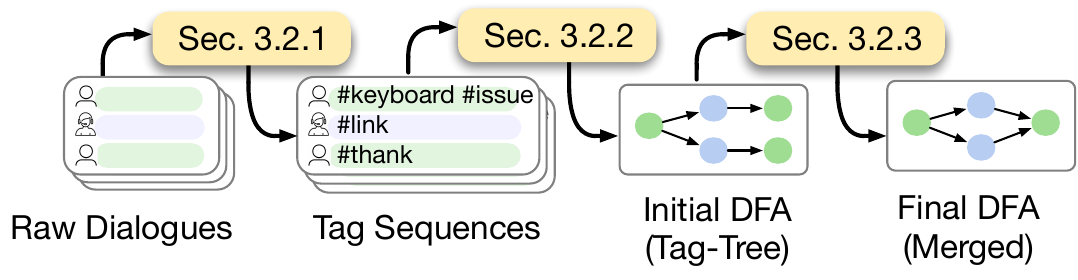}
    \vspace{-0.3cm}
    \caption{Outline for Section~\ref{sec:learn_dfa}. }
    \label{fig:outline}
\end{figure}

This section delves into the methodology of learning DFA from conversation sets, with a brief outline in Figure~\ref{fig:outline}. In Section ~\ref{sec:extract_tag}, we detail the process of deriving tag sequences from conversational data, employing LLMs to accurately identify and extract relevant tags. Following this, Section~\ref{sec:tree_build} focuses on the assembly of these tag sequences into a structured tree format, laying the groundwork for DFA construction. Lastly, in Section~\ref{sec:state_merge}, we introduce   the state-merging process within the tag tree, refining the DFA to succinctly represent the dynamics of conversational flows. 

\subsubsection{Extracting Tag Sequences from Conversations Using LLMs}
\label{sec:extract_tag}
In this subsection, we detail the method for converting a conversation into a sequence of tags, a crucial preprocessing step for constructing the DFA. The process hinges on the use of the state-of-the-art LLMs, such as GPT-4, which are adept at processing natural language and extracting key information like human beings. The primary task of the LLM in this context is to preprocess each sentence of a conversation and distill it into the most representative tags.

The extraction process is guided by specific prompt parameters to ensure that the resulting tags are both concise and relevant to the conversational content. These prompts are structured to instruct LLM adhering to two main criteria: a) \textbf{Brevity}: The LLM is instructed to use as few words as possible, ideally limiting the output to no more than three words per tag.
b) \textbf{Focus on Core Elements}: The summarization is directed to concentrate on actual events, issues, queries, or solutions present in the conversation. We include the full prompts in Appendix~\ref{sec:sup_prompts}.

By applying the prompts, the LLM processes each sentence of the conversation and outputs succinct, focused tags. These tags serve as the building blocks for constructing the DFA, encapsulating the key elements of the conversation. 

\subsubsection{Tree Construction with Tag Sequences}
\label{sec:tree_build}

The traditional approach to building a DFA often begins with the construction of a prefix tree, representing the sequences of alphabets that form valid strings in the language. In the case of conversation sets, however, we encounter two significant challenges that diverge from this traditional method:

\textbf{1. Tags in each utterance can appear in various orders}, which is unlike strings in traditional DFA models that follow a fixed sequence of alphabets. This flexible ordering presents a unique challenge in constructing a prefix tag tree. For example, in a standard DFA, the string ``ab" is distinctly represented by the sequence ``\texttt{a} $\rightarrow$ \texttt{b}". However, a sentence like ``My iPhone battery drains fast" in a conversation can be equivalently tagged as either ``\#\texttt{issues} $\rightarrow$ \#\texttt{battery}" or ``\#\texttt{battery} $\rightarrow$ \#\texttt{issues}". 
To efficiently manage this variability and reduce the size of the tag tree, our algorithm prioritizes tags based on their frequency of occurrence, preferring to place more frequent tags earlier in the sequence. 

\textbf{2. Each tag derived from an utterance is associated with the context of a specific round in the conversation}. For instance, the tag ``\#\texttt{address}'' may carry distinct semantic meanings in the user round compared to the system round. In one context, it might signify an address query, while in another, it serves to provide information. 
Therefore, when evaluating tag frequencies or constructing the tree, we need to focus exclusively on tags from the same conversational round. As illustrated in the left part of Figure~\ref{fig:tree-alg}, an important preprocessing step is to convert the original tag sequences to a special data structure $\Psi$ with ID-ed tag sets across different conversational rounds.

To address these challenges, we introduce the algorithm as depicted in Figure~\ref{fig:tree-alg}. It begins with constructing a tag tree from the initial state $q_0$ and  \textit{round 0}, detailed from \textit{Lines 2 to 8}. For each subsequent round, the newly constructed tree's root state becomes a child state of one of the leaf states from the previous round's tree. The selection of leaf states $Q$ for progression into the next round is guided by additional heuristic rules, which are elaborated in Appendix~\ref{sec:sup_exp_details}. The procedure for building a tree from a set of tags, along with associated dialogue IDs, is expanded upon in \textit{Lines 9 to 16}. The central strategy is recursively selecting the most frequent tag to expand the tree. 

Upon execution of Algorithm 1, we yield a tree-like structure resembling a DFA. However, during our final results demonstration in the right-bottom part of Figure~\ref{fig:tree-alg}, it becomes evident that duplicated sub-sequences, such as ``\texttt{\#link} $\rightarrow$ \texttt{\#thank}", exist. This redundancy presents an opportunity to merge these sequences, thereby reducing the overall size of the resultant DFA. 

\subsubsection{State Merging in Tag Tree}
\label{sec:state_merge}

In the process of constructing a DFA-tree from conversation sets, the variability in the tags generated by LLM for utterances with similar or identical contexts is a key challenge. For instance, tags like ``\texttt{\#subscription}" and ``\texttt{\#membership}" might represent the same conversation topic, yet linguistically, they are distinct. Merging such states based purely on linguistic similarities is impractical due to their semantic differences. However, in a conversational context, these tags often play similar roles and are connected to related sub-structures, such as common children-tags like ``\texttt{\#refund}" or ``\texttt{\#payment}", which are typically associated with financial aspects of subscriptions.

To address this, we introduce a similarity score \(\phi_{sim}(q, q')\) between two states $q$ and $q'$ as follows:
\[
\phi_{sim}(q, q') = \frac{ \sum_{t \in \mathcal{T}(q) \cap \mathcal{T}(q') } |\mathcal{I}(\delta(q, t))| \cdot |\mathcal{I}(\delta(q', t))|} {\sum_{t \in \mathcal{T}(q)} |\mathcal{I}(\delta(q, t))| \cdot \sum_{t \in \mathcal{T}(q')} |\mathcal{I}(\delta(q', t))|},
\]
where we define function $\mathcal{T}$ as the tag sets connected to q:
$$\mathcal{T}(q) = \left\{t \in \Sigma \mid \exists q^{\prime} \in Q, \delta(q, t)=q^{\prime}\right\}$$

Intuitively, $\phi_{sim}(q, q')$ calculates the ratio of common tags connected to both states $q$ and $q'$. The significance of each tag is considered by counting the number of dialogues related to the connected state, denoted by \(|\mathcal{I}(\delta(q, t))|\). This approach focuses on comparing the immediate (1-hop) children's states. Empirical evidence suggests that comparing direct children often suffices, though the formula can be extended to compare deeper sub-tree structures if necessary.

Once the similarity score is calculated, we set a threshold \(\lambda\) to determine which node pairs should be merged. Pairs with a similarity score exceeding this threshold, \(\phi_{sim}(q, q') > \lambda\), are considered for merging. This methodology allows us to consolidate similar conversational contexts, reducing redundancy and increasing the efficiency of the DFA-tree.

\begin{table*}[tb]
    \centering
    \caption{Results of dialogue generation quality across different base models and methods. This table reports the ``Win Rate'' over naive base models (GPT-4, GPT-3.5) regarding dialogue generation performance. For each method using in-context learning (\texttt{RandSamp}, \texttt{RAG}, \texttt{BM25}, \texttt{DFA-RAG}), we use 5 samples in the inference time. 
    For \texttt{FT-LLM}, we perform fine-tuning using the API provided by OpenAI with standard hyperparameters. Note that the API for fine-tuning GPT-4 is not available. }
    \vspace{0.2cm}
    \resizebox{0.97\linewidth}{!}{

\begin{tabular}{llccccccc}
\toprule
\multicolumn{1}{c}{\multirow{2}{*}{\textbf{Base LLM}}} & \multicolumn{1}{c}{\multirow{2}{*}{\textbf{Methods}}} & \multicolumn{6}{c}{\textbf{Domains}} & \multirow{2}{*}{\textbf{Average}} \\  \cline{3-8}
\multicolumn{1}{c}{} & \multicolumn{1}{c}{} & AmazonHelp & DeltaSupport & AskPlayStation & AirbnbHelp & NikeSupport & CambridgeInfo &  \\ \midrule 
\multirow{5}{*}{GPT-4} 
 & RandSamp & 69.1 & 84.1 & 57.9 & 78.3 & 45.3 & 67.0 & 66.9\\
 & BM25 & 67.3 & 81.5 & 63.8 & 77.1 & 59.8 & 63.0 & 68.7\\
 & RAG &  74.4 & 87.0 & \textbf{66.3} & 72.2 & 57.3 & 66.5 & 70.6 \\
 & FT-LLM & - & - & - & - & - & - & - \\ 
 &  \textbf{DFA-RAG (Ours)} & \textbf{78.0} & \textbf{89.9} & 65.9 & \textbf{80.9} & \textbf{62.6} & \textbf{68.5} & \textbf{74.3} \\ \midrule
 
\multirow{5}{*}{GPT-3.5} 
 & RandSamp & 70.2 & 83.6 & 61.3 & 69.5 & 58.9 & 57.9 & 66.9 \\
 & BM25 & 70.6 & 84.1 & 64.7 & 74.3 & 60.4 & 58.8 & 68.8 \\
 & RAG &  73.8 & 82.9 & 72.4 & 76.6 & 63.3 & 60.6 & 71.6 \\
 & FT-LLM & 69.7 & 64.6 & 71.7 & 66.1 & 56.8 & 56.1 & 64.2\\
 &  \textbf{DFA-RAG (Ours)} & \textbf{78.5} & \textbf{89.8} & \textbf{72.9} & \textbf{79.1} & \textbf{70.1} & \textbf{64.9} & \textbf{75.9} \\ 
 
 \bottomrule 
\end{tabular}
    
    }
    \label{tab:main}
\end{table*}

\subsection{Conversation Generation by DFA-RAG}
\label{sec:dfa_guide}

Once a DFA is constructed to model conversation sets, it can be integrated with a Large Language Model (LLM) to generate dynamic and coherent conversations as demonstrated in Figure~\ref{fig:teaser}. This process involves a cyclical interaction between the user's inputs, the DFA's navigational structure, and the LLM's response generation capabilities. Here are the concrete steps involved in this procedure:

\begin{enumerate}
    \setlength\itemsep{-0.3em}
    \item 
    \textbf{Tagging User Utterances}: When a user responds, their input is first processed by the LLM, which converts it into tags. This process mirrors the approach used for tagging the training samples, ensuring consistency in tagging the conversational content.
    \item \textbf{Navigating the DFA}: Based on the tags derived from the user's input, the system navigates through the DFA with a deterministic path to identify the corresponding state. %
    Given an input tag sequence $\{u_0, u_1, ..., u_T\}$, the state transition at step $t$ is denoted as $q_t = \delta(q_{t-1}, u_t)$. In cases where $\delta(q_{t-1}, u_t) = \varnothing$, indicating a mismatch, the navigation process concludes, and we proceed to retrieve dialogue IDs based on the last valid state, $q_{t-1}$. 
    
    \item \textbf{Accessing Dialogue IDs}: Each state $q$ in the DFA is associated with a collection of dialogue IDs $\mathcal{I}(q)$, which represent past conversations that have followed a similar pathway through the DFA. 
    \item \textbf{Compiling a Prompt for LLM}: From the dialogue IDs, the system samples $5$ examples and compiles them into a prompt structured for in-context learning as shown in Figure~\ref{fig:teaser}(c).
    \item \textbf{LLM Response Generation}: The LLM then generates a response based on the given prompt. 
    \item \textbf{Iterative Process}: The conversation continues with the user providing their next response based on the LLM's output. The process reverts to \textbf{STEP 1}, with the user's new response undergoing the same cycle of tagging, DFA navigation, and LLM response generation.
\end{enumerate}

This iterative process allows for a dynamic conversation flow, where each response is contextually informed and semantically relevant.

\section{Experimental Results}
\label{sec:exp}
\textbf{Datasets.} We carry out experiments with dialogue datasets in extensive domains. A high-level summary of the dataset statistics is provided in Table~\ref{tab:datasets}.

\begin{table}[ht]
    \centering
    \vspace{-0.4cm}
    \caption{Summary statistics of the datasets.}
    \vspace{0.2cm}
    \resizebox{0.97\linewidth}{!}{
    \begin{tabular}{l|llll}
    \toprule
    Domains & Dataset Size & Test Size & Avg. Rounds & Source\\
    \midrule
         AmazonHelp  & 50K & 1K & 6.4 & Tweet  \\
         DeltaSupport  & 20K & 1K & 4.4 & Tweet\\
         AskPlayStation & 10K & 500  & 3.8 & Tweet \\
         AirbnbHelp  & 3K & 100 & 4.2 & Tweet\\
         NikeSupport & 1K & 100 & 5.5 & Tweet\\
	CambridgeInfo  & 8K & 1k & 13.5 & MultiWOZ\\
    \bottomrule
    \end{tabular}}
    \label{tab:datasets}
\end{table}

The dataset described in the table consists of domain-specific collections, each serving distinct business communication needs. Notably, the first five datasets --- \texttt{AmazonHelp}, \texttt{DeltaSupport}, \texttt{AskPlayStation}, \texttt{AirbnbHelp}, and \texttt{NikeSupport}, are derived from Twitter interactions~\cite{tweet}, including a range of sizes from larger datasets containing over 50,000 dialogues (AmazonHelp and DeltaSupport) to smaller ones with 1,000 entries (NikeSupport). These datasets represent unique customer service interactions pertaining to their respective single business domains.

In contrast, \texttt{CambridgeInfo}, an alias for the Cambridge Information Center, encompasses a broader spectrum of services. It includes dialogues related to various booking services such as train, taxi, and hotel reservations. This dataset is sourced from MultiWOZ~\cite{budzianowski2018multiwoz}, and differs significantly from the Twitter-based datasets. MultiWOZ is known for its dense annotations, detailing the purpose and key information of each dialogue. This feature of MultiWOZ allows for a more nuanced analysis of dialogues in Section~\ref{sec:dst}, particularly in assessing the success rate of interactions.

\subsection{Generation Quality Evaluation}
\label{sec:gpt_eval}

This section aims to evaluate the quality of generated dialogues using the DFA-RAG framework. We compare performance on major pretrained LLMs as the backbone model in generating dialogues: \texttt{GPT-4}~\cite{openai2023gpt4}, \texttt{GPT-3.5}~\cite{ouyang2022training}. The concrete versions are \textit{gpt-4-1106-preview} and \textit{gpt-3.5-turbo-1106} respectively.

\textbf{Evaluation Technique}. We leverage GPT-4 for the primary evaluation, given its superior performance metrics. According to ~\citet{zheng2023judging}, GPT-4 demonstrates an 85\% agreement rate with human evaluators, signifying its reliability in assessing dialogue quality. We follow the methodology outlined in AlpacaEval~\cite{dubois2023alpacafarm} with similar prompts and metrics as detailed in Appendix~\ref{sec:sup_prompts}. The evaluation process involves GPT-4 comparing two dialogues to determine which one is closer to the ground truth. The dialogues completed directly by their respective LLMs serve as the target to be compared. The effectiveness of each method is quantified using the ``\textbf{Win Rate}" score, which reflects how often a dialogue surpasses the quality of the competitor.
For a qualitative comparison, we also present example dialogues with respective comments from the GPT-4 evaluator in Appendix~\ref{sec:sup_dialogue_demo}.

\textbf{Baselines}. We compare diverse strategies in selecting in-context learning (ICL) examples, recognizing the sensitivity of few-shot learning to sample selection: a)
\texttt{Random Sampling} involves randomly selecting examples from the training dataset; b)  \texttt{RAG} utilizes \textit{text-embedding-3-small} (from OpenAI) for generating sentence embeddings and focuses on retrieving training examples with utterances most similar to the test utterance; c) \texttt{BM25} is a classical sparse retrieval method used to find training examples with utterances closely matching those in the test scenarios. 
Additionally, we examine another baseline, ``\texttt{FT-LLM}" (Finetuned Large Language Model), which involves directly finetuning the LLM on the dialogues in the training set.

\textbf{Observations}.
Our observations from the evaluation in Table~\ref{tab:main} reveal some notable trends and insights: (a) the approach of fine-tuning (\texttt{FT-LLM}) showed underwhelming performance, which can be attributed to the limited volume of domain-specific training data, such as NikeSupport; (b)  \texttt{RAG} demonstrates stronger performance than $\texttt{RandSamp}$ and $\texttt{BM25}$, which is likely due to its more accurate retrieval of relevant samples; (c) DFA-RAG outperforms the best baseline by $~4\%$ at win-rate. Note that DFA-RAG not only facilitates the generation of high-quality dialogues but also provides a clear and interpretable structure, making it easier for humans to understand and analyze the dialogue flow.

\subsection{Dialogue Task Evaluation}
\label{sec:dst}

In the evaluation of dialogue systems, particularly within customer service domains, one key metric is the system's ability to resolve the user's inquiries and issues effectively. Our evaluation is performed on the MultiWOZ dataset (corresponds to \texttt{CambridgeInfo} domain in Section~\ref{sec:gpt_eval}), a widely recognized benchmark in the realm of task-oriented dialogue generation. 

In these dialogues, system responses are typically presented in a delexicalized form --- specific values for certain variables like time/address within the dialogue are replaced with placeholders. In line with this practice, our model also incorporates delexicalization by replacing specific examples in in-context learning with placeholder forms.

To quantitatively measure the performance of our model, we employ standardized evaluation metrics, namely ``\textbf{Inform}" and ``\textbf{Success}" rates. The ``Inform" rate measures the degree to which the system provides sufficient and relevant information to fulfill the user’s information needs, while the ``Success" rate evaluates the system's performance in terms of completing the user’s goal like booking a hotel. The goal is to understand the overall efficacy of the dialogue system in achieving the intended outcomes of the interaction. 

\begin{table}[ht]
    \centering
    \vspace{-0.4cm}
    \caption{Evaluation results on the task-oriented
dialogues.}
    \vspace{0.2cm}
    \resizebox{0.97\linewidth}{!}{
    \begin{tabular}{cl|ll}
    \toprule
    \begin{tabular}[x]{@{}c@{}}Ground Truth States \\ in Training?\end{tabular}
     & \textbf{Methods} & \textbf{Inform} & \textbf{Success} \\ \midrule

    \multirow{5}{*}{Yes} & HDSA & 87.9 & 79.4 \\
    & MarCo  & 94.5 & 87.2 \\
    & HDNO & 93.3 & 83.4  \\
    & GALAXY & 92.8 & 83.5 \\ 
    & KRLS & 93.1 & 83.7  \\
    \midrule
    \multirow{5}{*}{No} & AuGPT & 76.6 & 60.5 \\
    & MTTOD & 85.9& 76.5 \\
    & RSTOD & 83.5& 75.0 \\
    & RewardNet & 87.6 & 81.5 \\ 
    & TOATOD & 90.0 & 79.8 \\
    \midrule 
    No &  \textbf{DFA-RAG (Ours)} & 93.3 & 90.0\\  
    \bottomrule
    \end{tabular}}
    \label{tab:tod}
\end{table}

In Table~\ref{tab:tod}, we compared two sets of baseline methods: a) one branch of works (HDSA~\cite{chen2019hdsa}, MarCo~\cite{wang2020marco}, HDNO~\cite{wang2020hdno}, GALAXY~\cite{he2022galaxy}, KRLS~\cite{yu2023krls}) assumes known dialogue states for policy optimization strategy and response generation;  b) the other branch, also known as End-to-End models (AuGPT~\cite{kulhanek2021augpt}, MTTOD~\cite{lee2021mttod}, RSTOD~\cite{cholakov2022rstod}, RewardNet~\cite{feng2022rewardnet}, TOATOD~\cite{bang2023toatod}), rely solely on dialogue context. Our method falls into the latter category, focusing on the model's inherent ability to generate responses without using known dialogue states. Despite this, our approach has delivered promising results across both ``Inform'' and ``Success'' metrics, exhibiting performance comparable to methods relying on ground truth dialogue states.  This highlights the efficacy and adaptability of our model in handling a range of dialogue scenarios efficiently.

\begin{figure*}[htb]
    \centering
    \includegraphics[width=0.995\linewidth]{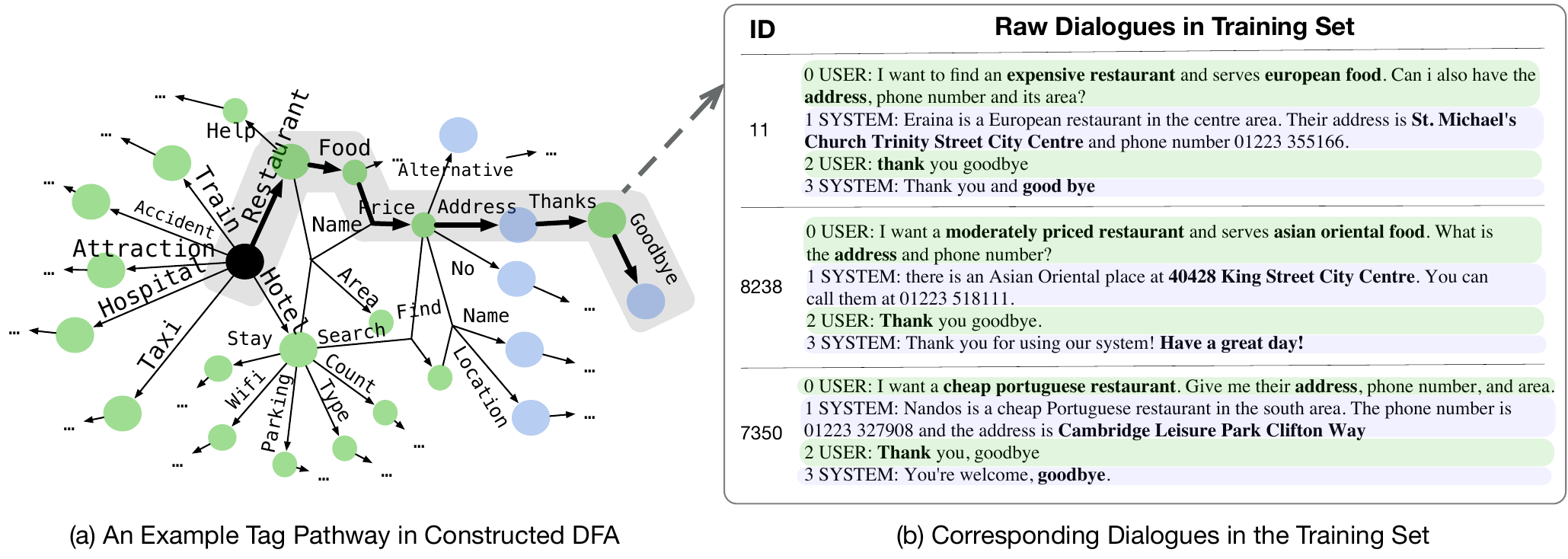}
    \caption{DFA Results for MultiWOZ. (a) This segment of the figure illustrates a portion of the constructed DFA. The black circle indicates the starting point of the automaton. Each green circle represents a ``user" state, while each blue circle denotes a ``system" state. The states are interconnected by arrows, each labeled with a tag. Note that some lines are interconnected (ex. lines correspond to``name'' and ``area''), it means that the relevant nodes are connected in both ways. 
(b) A specific path within the DFA is highlighted to demonstrate its correspondence with actual dialogues traversed. In these dialogues, elements associated with the tags are emphasized in bold.  }
    \label{fig:dfa-results}
\end{figure*}

\subsection{Constructed DFA Demonstrations} 
\label{sec:dfa_demo}

We display the (partial) DFA constructed from the MultiWOZ dataset in Figure~\ref{fig:dfa-results}. At the starting point of the automaton, we observe direct connections to several key domains such as ``\texttt{train}", ``\texttt{hotel}", and ``\texttt{taxi}". These domains represent the most frequently encountered topics in the MultiWOZ dataset.
As we delve into specific domains like restaurants and hotels, a notable observation is the recurrence of similar tags such as ``\texttt{price}", ``\texttt{name}", and ``\texttt{area}", which are common queries from users. Another intriguing aspect of the DFA is how the same state can lead to divergent system responses based on the user's input and the system's capabilities. For instance, if the system successfully resolves a user’s inquiry, as illustrated in the figure, it proceeds to provide specific details like the concrete address information. Conversely, another branch emerges when the system fails to find the desired target, signified by a ``no" tag. For readers' interest, we include more DFA demos in Appendix~\ref{sec:sup_dfa_demo}.
Overall, the DFA illustration captures the multifaceted nature of conversational flows within the MultiWOZ dataset and serves as a potent tool for understanding and improving domain-specific dialogue systems.

\section{Discussions and Future Work}
 \textbf{Handling Out-of-distribution Utterances}. 
In practice, it is common to encounter out-of-distribution (OOD) instances~\cite{yang2022openood, zhang2023openood,sun2023detecting, sun2023and, sun2024graph} that deviate from the pre-defined trajectory in the DFA constructed from training dialogues. Unlike existing research in OOD detection~\cite{du2024dream, ghosal2024overcome,ming2022exploit, ming2022delving}, which typically rejects user input when an OOD instance is detected, our approach leverages the dynamic capabilities of LLMs to respond to OOD utterances based on their best judgment—a feature absent in traditional rule-based dialogue systems.

 Formally, utterances are treated as  OOD in dialogues where $\delta(q_{t-1}, u_t) = \varnothing$, indicating no viable subsequent state (a mismatch scenario), the framework concludes the current navigation process and retrieves dialogue options based on the most recent valid state, $q_{t-1}$. This approach addresses situations where user inputs do not match any predefined pathway in the DFA, while still managing to navigate to the most relevant context. For example, if an unfamiliar user's inquiry occurs--such as ``booking \texttt{NBA} game tickets'', which might not directly correspond to a \texttt{NBA} tag within the DFA's structure, our framework adeptly guides the dialogue towards a related, more general context. This is achieved by reverting to the parent state $q_{t-1}$ which represents sports event ticket booking. This strategy ensures the DFA-RAG can maintain coherence and relevance in conversations, even when faced with partially matched or unexpected user inputs.

\textbf{Integration with External Modules.} The plug-and-play nature of our method facilitates seamless integration with existing services and algorithms. For instance, in scenarios where an input sentence's tags fall outside the predefined tracks of the DFA, our framework can dynamically interface with web search services or other external tools via callable APIs. This enables the LLM to access a broader range of information and resources to improve the quality of generated responses. 

\section{Related Work}
\label{sec:related}

\textbf{Structured Dialogue System.}
Traditional conversational agents operate on a fixed knowledge base, providing deterministic responses to user inputs from a pre-defined set of answers~\cite{abdul2015survey,bickmore2016improving,jain2018evaluating,medeiros2019towards,hussain2019survey,adamopoulou2020chatbots,janssen2020virtual,safi2020technical}. While reliable and consistent, they lack flexibility and struggle with unexpected queries. In contrast, ML-based conversational agents~\cite{openai2023gpt4,ouyang2022training}, represent a more adaptive approach but can be unpredictable.
Our proposed method combines the stability of traditional agents with the adaptability of ML-based systems, leveraging the strengths of both  to create a more dynamic and reliable dialogue system.

\textbf{Retrieval-Augmented Generation (RAG).}
RAG was originally introduced in ~\cite{lewis2020rag}, which established an embedding database containing accessible documents, enabling the retrieval of related information in generating responses. This approach addresses key limitations of LLMs, particularly in areas of knowledge updating and reasoning transparency~\cite{shuster2021retrieval, yasunaga2022retrieval, borgeaud2022improving, khattab2022demonstrate, cheng2022neural, wang2023knowledgpt, cheng2023lift}. 
Building upon the naive RAG, more sophisticated variants have been developed. These models incorporate complex components like structured data sources. For instance, RET-LLM ~\cite{modarressi2023ret} constructs a personalized knowledge graph memory, extracting relation triples to enhance response generation. Similarly, SUGRE~\cite{kang2023knowledge} utilizes Graph Neural Networks (GNN) to embed relevant sub-graphs retrieved from the knowledge graph. KnowledgeGPT~\cite{wang2023knowledgpt} generates search queries for Knowledge Bases (KB) in code format and includes predefined KB operation functions.
Our algorithm can be also viewed as an extension of the RAG concept, which encodes historical dialogues into DFA and retrieves similar dialogues based on it. Our approach marks a pioneering step in conversational agent research.

\textbf{Task-oriented Dialogue System.}
In this paper, we aim to enhance the reliability of LLMs for domain-specific dialogue generation. A critical subset involves task-oriented dialogue (TOD) systems, which are designed to accomplish specific tasks through interactive communication. A notable line of  methods involves leveraging dialogue state annotations to train policy-based dialogue generation systems~\cite{chen2019hdsa, ramachandran2021caspi, tseng2021joust, wang2020marco, wang2020hdno, he2022galaxy, yu2023krls}.
Alternatively, some approaches have simplified the design of TOD systems to the end-to-end (E2E)~\cite{kulhanek2021augpt, lee2021mttod, su2022multi, cholakov2022efficient, cholakov2022rstod, feng2022rewardnet, bang2023toatod}.
Comparing all existing works, our DFA-RAG framework introduces a  distinct approach without relying on traditional gradient-based training. This approach offers simplicity and adaptability, allowing for easy and efficient adjustments to new scenarios in a plug-and-play manner.

\textbf{Semantic Router.}
The Semantic Router~\cite{horsey2024semantic, azhar2024beyond, avila2024semantic, sisodia2024stop, hingane2024llm} represents a recent advancement in LLM technology, particularly in augmenting the capabilities of chatbots and AI assistants through the development of a predefined decision-making layer. This router directs the decision trajectory of LLMs by evaluating semantic similarity in each branch and selecting the closest match. Typically, similarity is measured using cosine distance between the embedding vector of inquiries and the potential branches. In this paper, DFA-RAG can be seen as an evolution of the semantic router, offering several key advantages:
a) DFA-RAG navigates the decision-making layer using tags instead of embedding comparison, which enhances interpretability and computational efficiency.
b) Unlike decision-making layers predefined by humans\footnote{See Aurelio AI \url{https://www.aurelio.ai/}.}, the DFA structure within our framework is learnale from historical training data, making it more adaptable and scalable for real-world applications.

\section{Conclusion}
\label{sec:conclusion}

This paper proposes the DFA-augmented Large Language Model (DFA-RAG), integrating the adaptability of large language models with the structured approach of Definite Finite Automata. The DFA structure is learnable from the dialogues in the training set.  The whole framework ensures reliable, contextually appropriate responses, addressing key limitations in current LLM applications, particularly in specialized areas like customer service.  Extensive experiments are conducted to validate DFA-RAG’s effectiveness in generating pertinent dialogue content. 
We hope this pioneering research can pave the way for more controlled and predictable AI conversational agents without sacrificing the dynamic nature of LLMs. 


\section*{Impact Statement}
The DFA-RAG framework has the potential to revolutionize how conversational agents are deployed across sectors, including healthcare, customer service, and education, by providing more accurate, context-aware, and ethically aligned interactions. This advancement could lead to broader accessibility of digital services, enhanced user experiences, and greater trust in AI technologies. Beyond the mentioned societal consequences, we do not anticipate any potentially harmful consequences to our work. 

\section*{Acknowledgment}
Hu is supported by the Wisconsin Alumni Research Foundation, and by the National Institute Of Biomedical Imaging And Bioengineering of the National Institutes of Health under Award Number R01EB033782. The content is solely the responsibility of the authors and does not necessarily represent the official views of the National Institutes of Health.

\bibliography{main}
\bibliographystyle{icml2024}

\newpage
\appendix
\onecolumn
\section{Prompts}
\label{sec:sup_prompts}

\paragraph{Prompts used for extracting tags.} This prompt template is used in Section~\ref{sec:extract_tag}  to make sure the resulting tags are both concise and relevant to the conversational content.

\begin{lstlisting}
# Task Description

You are helping me compress the following dialog with customer service into the following form:

<id> <User/System>: <compressed phrase>

You will have to follow several principles:
1. Please use words as few as possible, ideally no more than 3 words.
2. The summarization needs to focus on the actual events/issues/queries/solutions.


# Example

Input:
"0 User: What is going on with my keyboard... fix it"
Output:
"0 User: #keyboard #issue"

\end{lstlisting}

\paragraph{Prompts used for generating response.} This prompt template is used in Section~\ref{sec:dfa_guide} to generate the response given the example dialogues.

\begin{lstlisting}
# Task Description
You are a helpful service agent. Please help me fill in the system response in a dialogue.
Please note that key information is encoded in the dialogue.

The dialogue is with the format:

[ID] [USER/SYSTEM]: [UTTERANCE]

Here is a list of related example dialogues you can use for reference.

{examples}

# Remarks:

1. Please directly generate the completed dialogue according to the format in the example.
2. (**IMPORTANT**) Please make sure the generated utterance ID is consistent with the original input!

\end{lstlisting}

\paragraph{Prompts used for evaluation.} This prompt template is used in Section~\ref{sec:gpt_eval} to evaluate the response quality and win rate calculation. This prompt is modified from AlpacaEval~\cite{dubois2023alpacafarm}.

\begin{lstlisting}

I'll provide you with task prompts given to these models and their corresponding outputs. 
    Your task is to assess these responses, and select the model that produces the output that is most smooth and consistent with the ground truth dialogue. 
    Please note that it is very important for model to provide response in a **similar style and content**.
    
    ## Instruction
    
    {{
        "instruction": """Please act as a helpful customer service agent and complete the following dialogue: """,
        "input": 
        """
            {task_input}
        """,
        "ground truth answer": """
            {raw_diag}
        """
    """
    }}
    
    ## Model Outputs
    
    Here are the unordered outputs from the models. Each output is associated with a specific model, identified by a unique model identifier.
    
    {{
        {{
            "model_identifier": "m",
            "output": """
    {pred_cmp_diag}
            """
        }},
        {{
            "model_identifier": "M",
            "output": """
    {pred_diag}
            """
        }}
    }}
    
    ## Task
    
    Evaluate the models based on the quality and relevance of their outputs, and select the model that generated the best output. 
    Answer by first providing a concise explanation and then end your answer by providing the model identifier of the best output. 
    We will use the last character of your output `output[-1]` as the name of the best model, so make sure you finish with the token of the model identifiers and nothing else: `m` or `M` (no quotes, no dots, no backticks, no new lines, ...). 
    For example:
    
    ### Concise explanation
    ...some text...
    
    ### Which is best, m or M?
    M
    
    Now is your turn.
    
    ## Your answer: "Concise explanation" followed by "Which is best, m or M?"

\end{lstlisting}

\section{Experiment Details}
\label{sec:sup_exp_details}

In this section, we delve into the specifics of implementing the DFA learning process from training data, as outlined in Section~\ref{sec:learn_dfa}, and the intricacies of conversation generation, discussed in Section~\ref{sec:dfa_guide}.

\textbf{A threshold for determining the ``splittable'' states}. 
Figure~\ref{fig:tree-alg} illustrates the algorithm for constructing the DFA-tree. However, the practical application of this algorithm involves several nuanced considerations. One key aspect is the handling of nodes/states that correspond to only a limited number of dialogues in the training dataset. Constructing a sub-tree under such nodes/states may not yield significant insights due to their limited number of tracked dialogues. Therefore, the tree-building function, as depicted in Figure~\ref{fig:tree-alg}, is only pursued if the number of tracked dialogue IDs at a node/state exceeds a certain threshold: \(|\mathcal{I}(q)| > \tau\). For our experiments, we have set this threshold, \(\tau\), to 5.

\textbf{State-merging threshold}. In Section~\ref{sec:state_merge}, we illustrate that once the similarity score is calculated, we set a threshold \(\lambda\) to determine which node pairs should be merged. In practice, we set $\lambda = 0.1$ and by observation, ~$5\%$ of nodes get merged with the threshold set. 

\textbf{Examples limits in In-Context Learning (ICL)}. 
In alignment with the practice in in-context learning ~\cite{li2023finding, dong2022survey}, we limit the maximum number of examples used in in-context learning to 5. This means that if the number of tracked dialogue IDs for a state, denoted as \(|\mathcal{I}(q)|\), exceeds 5, we randomly select 5 dialogues from \(|\mathcal{I}(q)|\) for further processing.

\vspace{-0.3cm}
\section{Dialogue Generation Results Demonstrations}
\label{sec:sup_dialogue_demo}

In this section, we conduct a qualitative analysis of dialogue generation outcomes using the baseline algorithms outlined in Table~\ref{tab:main}. The dialogues are derived from the \texttt{AirbnbHelp} dataset, with all baseline models utilizing GPT-3.5 as the underlying large language model (LLM) for generation. Our objective is to compare each model's output with the ground truth dialogue from the dataset, aiming for generated dialogues that closely mirror the original interactions. Additionally, we include insights from a GPT-4 evaluator to shed light on the relative performance of each model.

\subsection{Ground Truth Dialogues}
\begin{tcolorbox}[colback=gray!5!white]
\begin{enumerate}
    \item \texttt{User}: I just received a mail asking me to give feedback on a trip that I didn't make. Error at your end or has my account been hacked?
    \item \texttt{System}: Hey Nick, thank you for bringing this to our attention. We've escalated a case to the appropriate team and a Case Manager will follow up with you for further assistance. We appreciate your patience in the meantime.
    \item \texttt{User}: I presume you got my message via the Help Center contact form?
    
    \item \texttt{System}: Yes, we've also noticed that your case manager has just sent you an email with all the instructions to recover your account. Please let us know if you need further assistance. Thanks.
    
    \item\texttt{User}: Well, I changed my password, and my account seems to be working. I canceled the trip and it said I would get a refund, but whoever made the trip didn't use my credit card, so I didn't pay!
    
    \item \texttt{System}: Hey Nick, we've looked into it and everything looks good. You shouldn't have to upload your ID again, as we see your account is still fully verified. Please let us know if there's anything else we can do for you. Thank you.
\end{enumerate}

\end{tcolorbox}

\subsection{GPT-3.5 Generation Results}
Below, we display the dialogue completions generated by GPT-3.5, without using any additional prompts:
\begin{tcolorbox}[colback=gray!5!white]
\begin{enumerate}
\item \texttt{User}: I just received a mail asking me to give feedback on a trip that I didn't make. Error at your end or has my account been hacked?
\item \texttt{System}: It seems like there might have been some misunderstanding. I'll need to investigate this further. \textcolor{red}{Could you please provide me with the email address that received the feedback request?}  \vspace{0.2cm} \\ 
\footnotesize \textcolor{red}{(\textbf{Comment from GPT-4}: It asks for the email address from the user, although presumably, the system should already have this information since the user received an unwanted email from the service.)}
\item \texttt{User}: I presume you got my message via the Help Center contact form?
\item \texttt{System}: Yes, I received your message through the Help Center contact form. Thank you for reaching out to us that way.
\item \texttt{User}: Well, I changed my password, and my account seems to be working. I canceled the trip and it said I would get a refund, but whoever made the trip didn't use my credit card, so I didn't pay!
\item \texttt{System}: I'm glad to hear that you changed your password and that your account is working. Our team will investigate the unauthorized trip and ensure that you are not charged for it. Rest assured that we will take care of the refund process for you.
\end{enumerate}

\end{tcolorbox}
These results, while informative, reveal certain limitations in capturing the nuances of the original conversation, as will be discussed in the evaluator's comments.

\subsection{DFA-RAG Generation Results (Ours)}

We next present results from the DFA-RAG approach. The GPT-4 evaluator offers more favorable comments for these outputs, highlighting an improved alignment with the ground truth dialogue:

\begin{tcolorbox}[colback=gray!5!white]
\begin{enumerate}
\item \texttt{User}: I just received a mail asking me to give feedback on a trip that I didn't make. Error at your end or has my account been hacked?
\item \texttt{System}: Hi there, it sounds like there may have been some confusion with your account. We'll investigate this further and get back to you with more information. Thank you for bringing this to our attention. 
\item \texttt{User}: I presume you got my message via the Help Center contact form?
\item \texttt{System}: Our team is currently reviewing your message and will provide you with an update as soon as possible. Thank you for your understanding.
\item \texttt{User}: Well, I changed my password, and my account seems to be working. I canceled the trip and it said I would get a refund, but whoever made the trip didn't use my credit card, so I didn't pay!
\item \texttt{System}: I'll ensure that your account is thoroughly checked to confirm its security and functionality. We appreciate your proactive steps, and we'll keep you updated on any developments. \vspace{0.4cm}\\
\footnotesize \textcolor{codegreen}{(\textbf{Comment from GPT-4}: This model does not ask for additional information that it should already have and instead focuses on confirming that the issue will be investigated and the user will be updated. This is closer to the ``ground truth'' response mentioned in the instruction, where the system acknowledges the receipt of the message and provides assurance without requesting information it should already possess. In addition, this model uses language that is more in line with the professional customer service manner shown in the ground truth dialogue, such as thanking the user multiple times and indicating appreciation of the user's proactive measures.)}
\end{enumerate}
\end{tcolorbox}
\subsection{RAG-Based Generation
Results}
Results from the RAG (Retrieval-Augmented Generation) model are also presented. This model retrieves dialogues that closely match the context in the embedding space, using these selections as prompts for generation. However, the generated responses often inaccurately focus on requesting the user's email, failing to directly address the user's initial request:

\begin{tcolorbox}[colback=gray!5!white]
\begin{enumerate}
\item \texttt{User}: I just received a mail asking me to give feedback on a trip that I didn't make. Error at your end or has my account been hacked?
\item \texttt{System}: Hi there, we're sorry to hear about this. Can you please DM us with your account email so we can look into this for you and provide further assistance? Thank you.
\item \texttt{User}: I presume you got my message via the Help Center contact form?
\item \texttt{System}: Thanks for confirming that. We'll review your account and ensure everything is in order. If you have any further concerns, please don't hesitate to reach out to us.
\item \texttt{User}: Well, I changed my password, and my account seems to be working. I canceled the trip and it said I would get a refund, but whoever made the trip didn't use my credit card, so I didn't pay!
\item \texttt{System}: We understand your concern and will definitely look into it for you. Please DM us your email address so we can verify your account and address any potential security issues. Thank you.
\vspace{0.4cm}\\
\footnotesize \textcolor{red}{(\textbf{Comment from GPT-4}: It did not address the final user query. Furthermore, the model started well by asking for a Direct Message (DM) for privacy, it did not fully reassure the user regarding the data breach or provide clear next steps concerning the unauthorized trip.)}
\end{enumerate}
\end{tcolorbox}

\subsection{RandSample Generation Results}
Finally, we discuss outcomes from the \texttt{RandSample} method, which randomly selects dialogues from the training dataset for prompt generation. This approach sometimes leads to generated dialogues that contain hallucinated content, such as mismatched usernames, illustrating the pitfalls of random selection:
\begin{tcolorbox}[colback=gray!5!white]
\begin{enumerate}
\item \texttt{User}: I just received a mail asking me to give feedback on a trip that I didn't make. Error at your end or has my account been hacked?
\item \texttt{System}: We appreciate you reaching out, \textcolor{red}{Simona}. It seems there may have been a mix-up. Please DM us your email address so we can look into this further for you.
\item \texttt{User}: I presume you got my message via the Help Center contact form?
\item \texttt{System}: We understand your concern, \textcolor{red}{Simona}. We assure you that we will investigate this and get back to you with an update.
\item \texttt{User}: Well, I changed my password, and my account seems to be working. I canceled the trip and it said I would get a refund, but whoever made the trip didn't use my credit card, so I didn't pay!
\item \texttt{System}: We'll definitely look into this for you, \textcolor{red}{Simona}. Please DM us your email address so we can check the status of your account and the refund for the canceled trip. Thank you.
\vspace{0.4cm}\\
\footnotesize \textcolor{red}{(\textbf{Comment from GPT-4}: This model consistently addresses the user by the name ``Simona," which does not align with the ground truth dialogue where the user is referred to as Nick. )}
\end{enumerate}
\end{tcolorbox}

Each of these methodologies offers unique insights into the challenges and opportunities present in dialogue generation. Through the lens of the GPT-4 evaluator's comments, we gain a nuanced understanding of how each model performs relative to the complexities of real-world conversation dynamics.
\section{More DFA Results Demonstrations}
\label{sec:sup_dfa_demo}

\begin{figure*}
    \centering
    \includegraphics[width=0.99\linewidth]{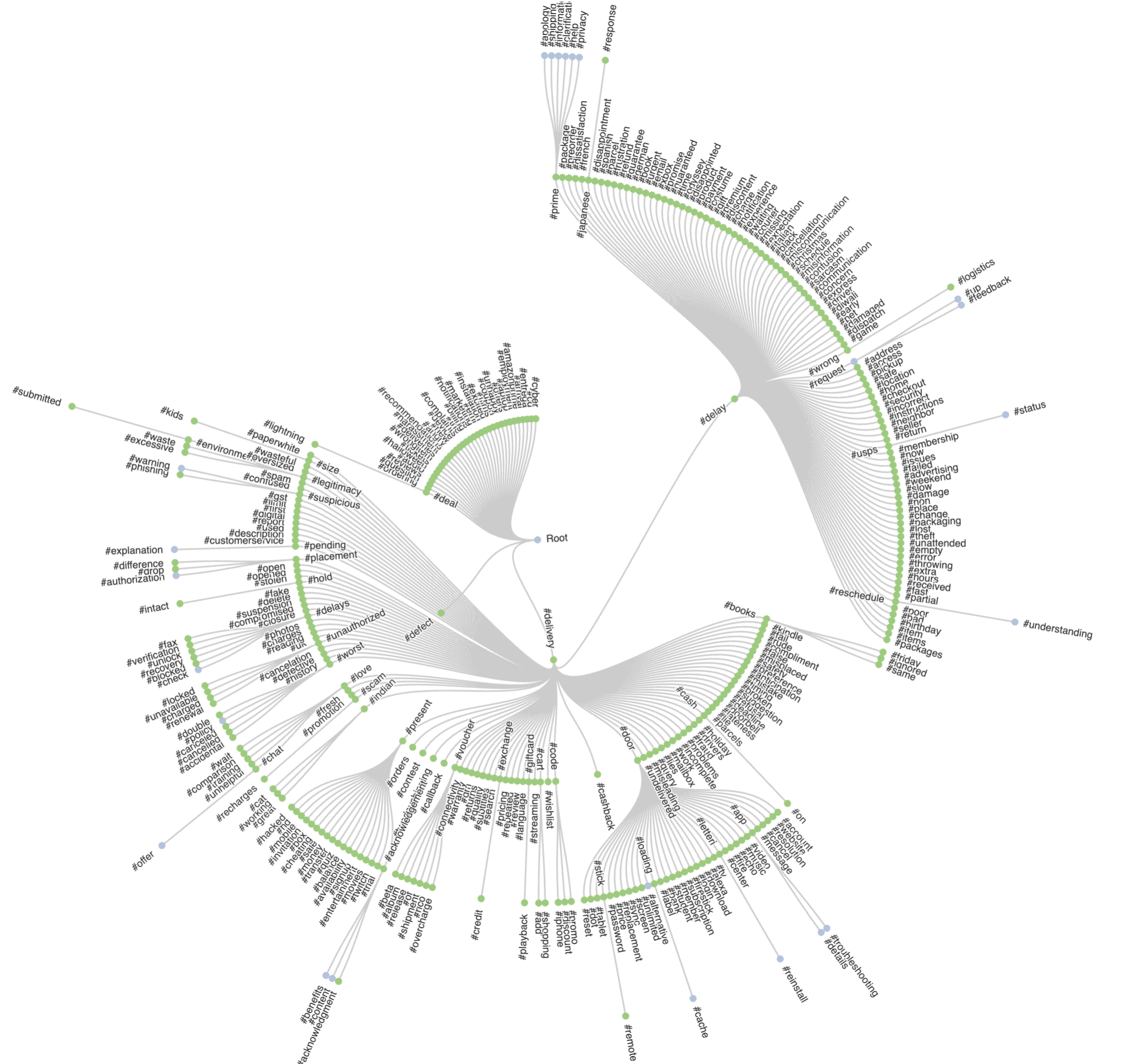}
    \caption{DFA Results Demonstration for \texttt{AmazonHelp}.}
    \label{fig:dfa-amazon}
\end{figure*}

\begin{figure*}
    \centering
    \includegraphics[width=0.99\linewidth]{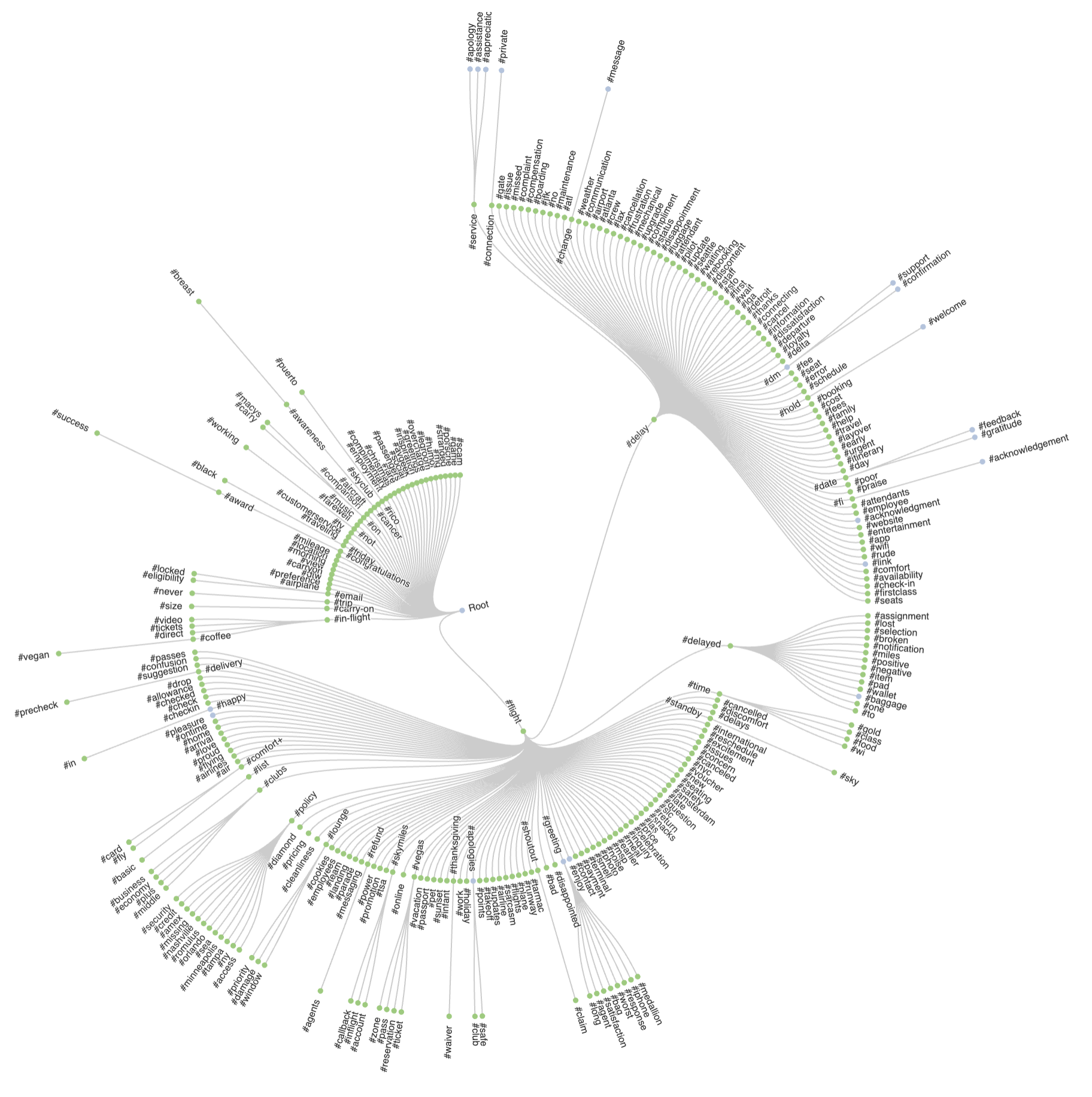}
    \caption{DFA Results Demonstration for \texttt{DeltaSupport}.}
    \label{fig:dfa-delta}
\end{figure*}

\begin{figure*}
    \centering
    \includegraphics[width=0.99\linewidth]{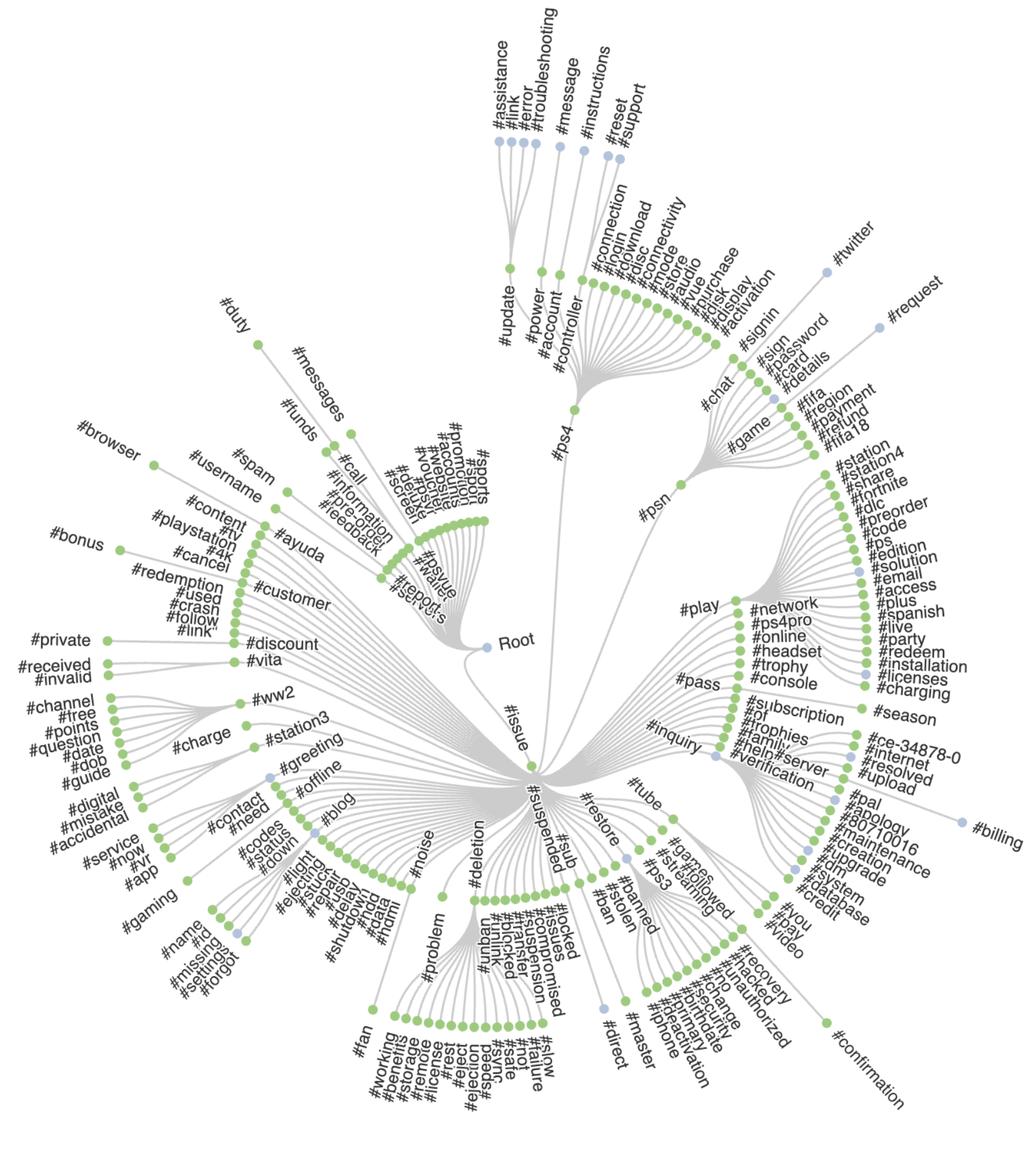}
    \caption{DFA Results Demonstration for \texttt{AskPlayStation}.}
    \label{fig:dfa-ps}
\end{figure*}

\begin{figure*}
    \centering
    \includegraphics[width=0.99\linewidth]{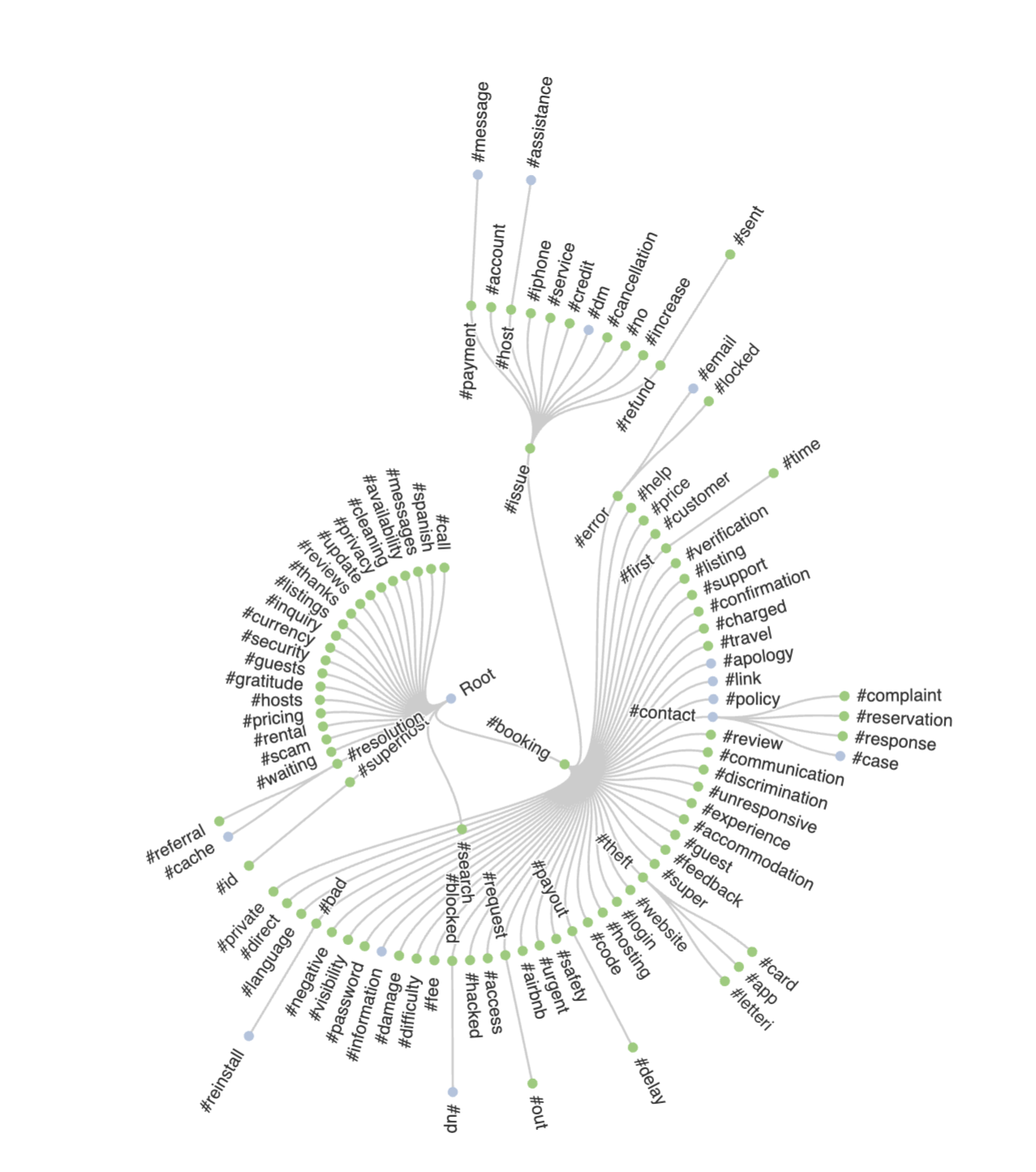}
    \caption{DFA Results Demonstration for \texttt{AirbnbHelp}.}
    \label{fig:dfa-airbnb}
\end{figure*}

\begin{figure*}
    \centering
    \includegraphics[width=0.99\linewidth]{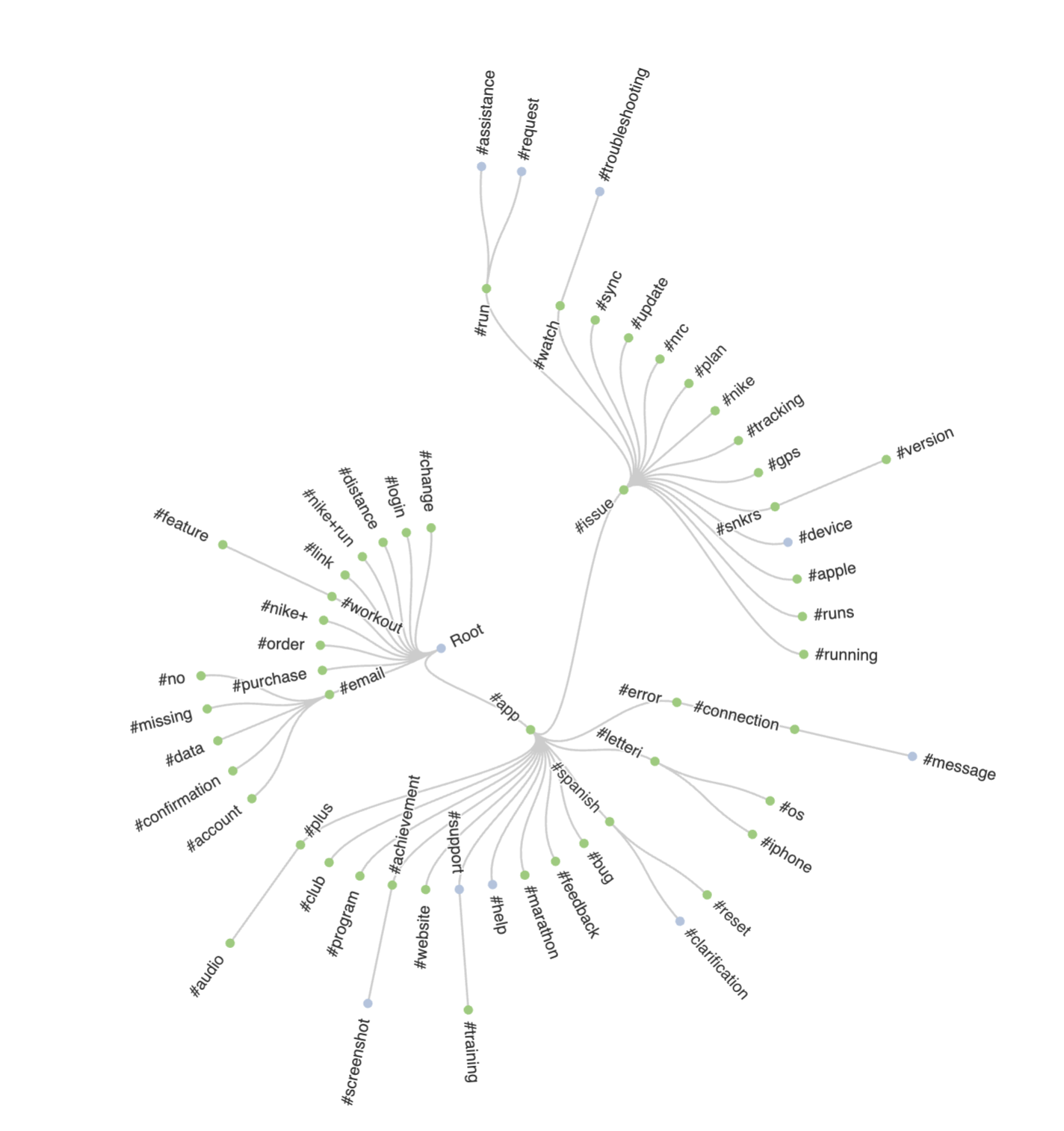}
    \caption{DFA Results Demonstration for \texttt{NikeSupoort}.}
    \label{fig:dfa-nike}
\end{figure*}

\begin{figure*}
    \centering
    \includegraphics[width=0.99\linewidth]{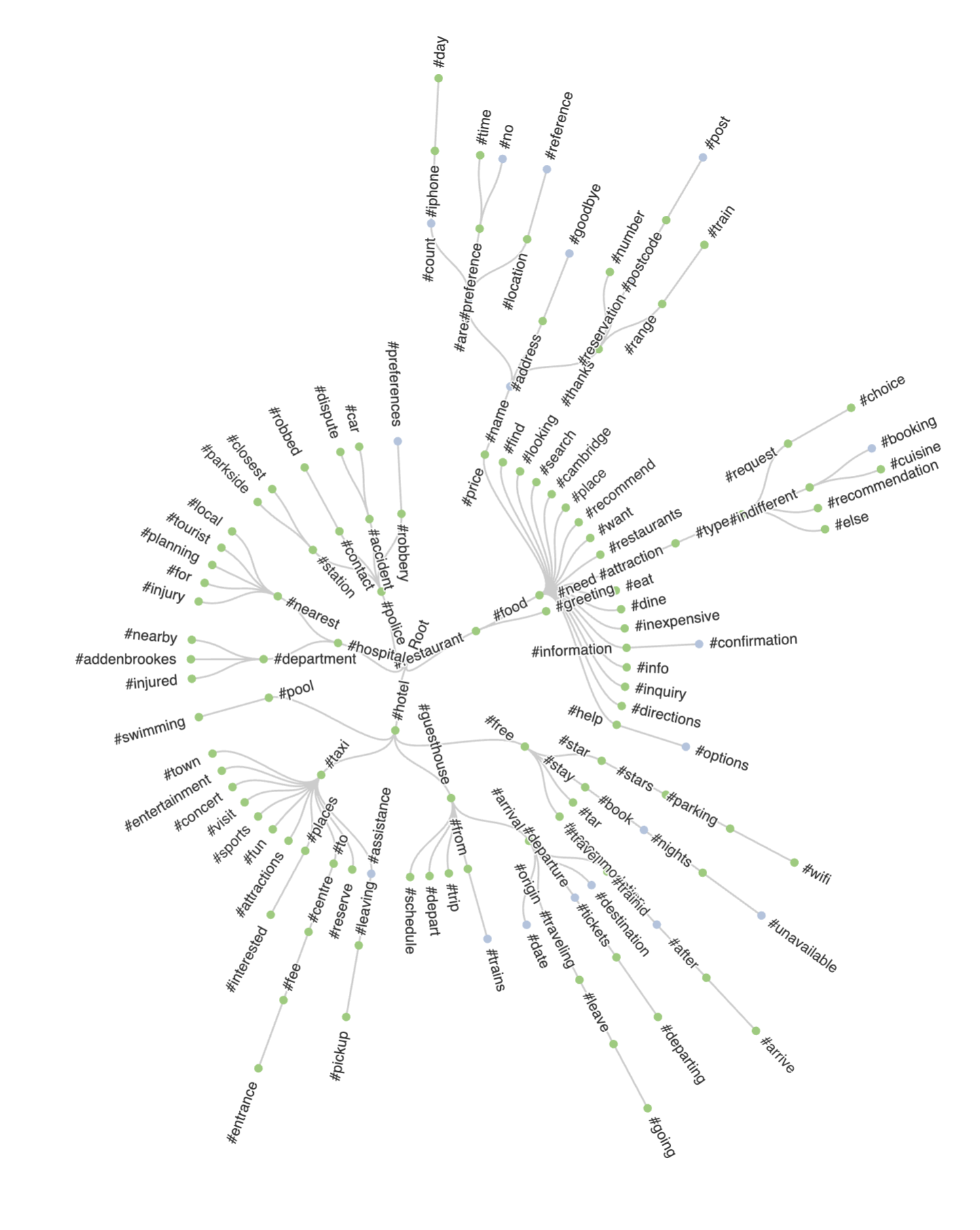}
    \caption{DFA Results Demonstration for \texttt{CambridgeInfo}/\texttt{MultiWOZ}. We extend the depth limitation to eight in the demo due to the complicated branching in this demo. }
    \label{fig:dfa-mwz}
\end{figure*}

 We include the constructed DFA for each dataset in Figure~\ref{fig:dfa-amazon}, Figure~\ref{fig:dfa-delta}, Figure~\ref{fig:dfa-ps}, Figure~\ref{fig:dfa-airbnb}, Figure~\ref{fig:dfa-nike}, Figure~\ref{fig:dfa-mwz}, we have constrained the maximum depth to four levels due to spatial constraints, and have chosen to display only those nodes that correspond to a minimum of ten dialogues in the training set, as indicated by $|\mathcal{I}(q)| > 10$. It's important to note that, owing to the limitations of our tree-like visualization tool, the `looping' structures typical of DFAs, characterized by nodes with multiple incoming edges, are not depicted here. Additionally, for clarity in this visualization, tags are shown attached to the nodes; however, in the actual DFA model, these tags are associated with the edges leading to each state/node. This visualization is intended to provide an overarching view of the DFA's structure. Readers should be mindful of these distinctions to accurately understand and interpret the DFA results.

\end{document}